\documentclass[letterpaper]{article}

\usepackage{aaai2027}
\usepackage[hyphens]{url}
\usepackage{graphicx}
\usepackage{natbib}
\usepackage{caption}
\usepackage{algorithm}
\usepackage{algorithmic}

\usepackage{newfloat}
\usepackage{listings}
\DeclareCaptionStyle{ruled}{labelfont=normalfont,labelsep=colon,strut=off}
\floatstyle{ruled}
\newfloat{listing}{tb}{lst}{}
\floatname{listing}{Listing}

\usepackage{booktabs}

\usepackage{subcaption}
\usepackage{amsfonts}
\usepackage{amsmath}
\usepackage{mathtools}
\usepackage{multirow}
\usepackage{siunitx}
\usepackage{hhline}
\usepackage{makecell}
\usepackage[dvipsnames]{xcolor}

\definecolor{yamlkey}{RGB}{0,128,0}
\definecolor{yamlstring}{RGB}{186,33,33}
\definecolor{yamlcomment}{RGB}{61,122,122}
\lstdefinelanguage{yaml}{
    keywords={routing, scoringRules, condition, tenants, targetPredictorName, geographies, schemas, shadowRules, targetPredictorNames},
    keywordstyle=\color{yamlkey}\bfseries,
    stringstyle=\color{yamlstring},
    morestring=[b]",
    commentstyle=\color{yamlcomment}\itshape,
    morecomment=[l]{\#},
    sensitive=true,
}

\title{MUSE: Multi-Tenant Model Serving with Seamless Model Updates}
\author{
    Cláudio Correia\equalcontrib,
    Lucas Martins\equalcontrib,
    Alberto E. A. Ferreira,
    Miguel P. Bento,
    Sofia Guerreiro,\\
    Ricardo Ribeiro Pereira,
    Ana Sofia Gomes,
    Jacopo Bono,
    Hugo Ferreira,
    Pedro Bizarro
}
\affiliations{
    Feedzai, Portugal\\
    \{firstName\}.\{lastName\}@feedzai.com
}

\begin{document}

\maketitle

\begin{abstract}
    In binary classification systems, decision thresholds translate model scores into actions. Choosing suitable thresholds relies on the specific distribution of the underlying model scores but also on the specific business decisions of each client using that model. However, retraining models inevitably shifts score distributions, invalidating existing thresholds. In multi-tenant \textit{Score-as-a-Service} environments, where decision boundaries reside in client-managed infrastructure, this creates a severe bottleneck: recalibration requires coordinating threshold updates across hundreds of clients, consuming excessive human hours and leading to model stagnation. We introduce MUSE, a model serving framework that enables seamless model updates by decoupling model scores from client decision boundaries. Designed for multi-tenancy, MUSE optimizes infrastructure re-use by sharing models via dynamic intent-based routing, combined with a two-level score transformation that maps model outputs to a stable, reference distribution. Deployed at scale by Feedzai, MUSE processes over 4 thousand events per second, and over 55 billion events in the last 12 months, across several dozens of tenants, while maintaining high-availability and low-latency guarantees.
    By reducing model lead time from weeks to minutes, MUSE promotes model resilience against shifting attacks, saving millions of dollars in fraud losses and operational costs.
\end{abstract}

\begin{links}
    \link{Code}{https://github.com/feedzai/muse-quantile-study}
    \par\textbf{Note} --- This is an extended version of the paper accepted at IAAI-27 (AAAI 2027). It is identical to the published version except for the appendices.
\end{links}

\begin{figure}[h]
    \centering
    \includegraphics[width=0.99\columnwidth]{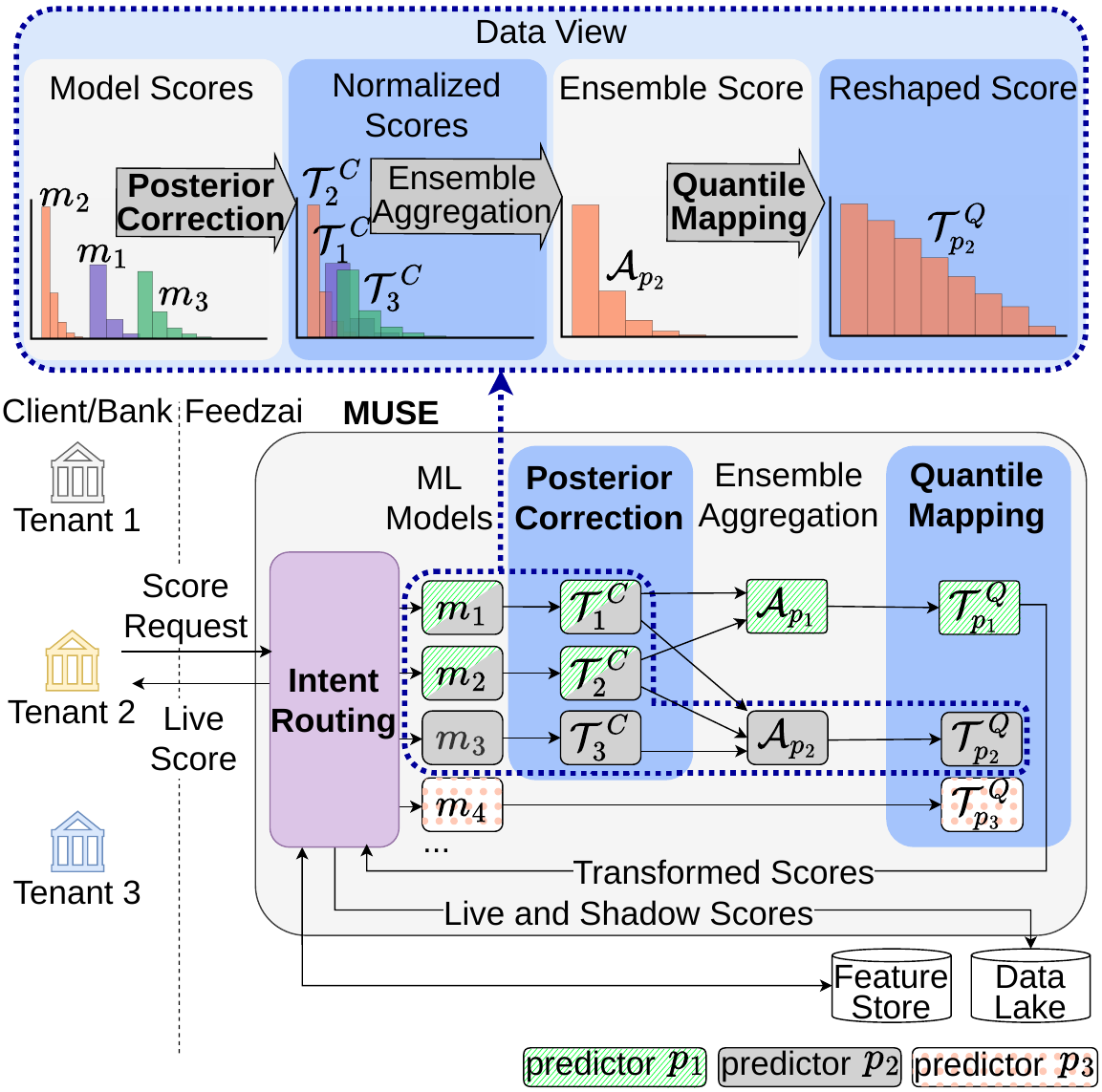}
    \caption{MUSE overview with predictors $p_1$ and $p_2$ (both ensembles), and predictor $p_3$, a standalone model. The top section details the scoring data pipeline for $p_2$.}
    \label{fig:diagram_1}
\end{figure}

\section{Introduction}
\label{sec:intro}

Machine Learning (ML) scoring systems are the backbone of modern large-scale decision pipelines, including fraud detection, credit underwriting, and healthcare~\cite{BrancoAGAAB20, creditscore, healthcare}. In these, models do not make decisions in isolation; they produce risk scores that are converted into actions (e.g., \emph{block}/\emph{allow}/\emph{review}) via decision thresholds. These thresholds are carefully selected by each financial institution to satisfy specific operational constraints, such as the daily capacity of fraud analysts or regulatory false-positive limits, intertwining the business logic with the score distribution of the deployed model.

This coupling creates a conflict in the model lifecycle. In high-stakes, adversarial use cases such as fraud detection, models must be updated often to remain effective against shifting fraud patterns and concept drift~\cite{conde2022, pinto2019, stripe_radar}. However, a retrained model inherently produces a different score distribution. For example, a threshold of $0.9$ in model v1 might capture $1\%$ of the riskiest transactions, but in model v2 it might capture $5\%$ (flooding analysts with false alarms) or $0.1\%$ (allowing fraud to pass undetected causing large monetary losses), both unacceptable outcomes in production.

This issue is amplified in multi-tenant \textit{Score-as-a-Service} setups, where a single provider manages and updates the ML models, but the decision boundaries (thresholds and rules) reside within the realm and infrastructure of individual clients\footnote{This paper uses the terms "client" and "tenant" interchangeably.}. In other words, the service provider only receives the input data required for inference, having no visibility nor authority to modify the client's internal rule thresholds.
Existing solutions to address score distribution shift fall short in this setting. Probability calibration~\cite{zadrozny2002transforming} generally requires extensive labeled data to map scores to probabilities, making it infeasible for fraud detection, where labels are delayed, sparse, or non-existent (e.g., in new deployments). Conversely, the traditional industry approach attempt to maintain the alert rate by manual threshold tuning at the tenant level~\cite{stripe_radar, kount_omniscore}, which does not scale. As the client base grows, the operational overhead of coordinating simultaneous threshold updates leads to ``model stagnation''~\cite{sculley2015hidden}, where inferior legacy models linger in production as the cost of upgrading is too high. In fraud detection, a highly adversarial domain, stagnation is unaffordable; continuous updates are fundamental to counter constantly evolving attacks ~\cite{qiao2025online}.

To address these challenges, we introduce MUSE, a model serving framework to enable seamless model updates that require no client intervention or awareness in multi-tenant \textit{Score-as-a-Service} settings, and under strict low-latency and high-availability requirements.
Depicted in Figure~\ref{fig:diagram_1}, MUSE abstracts the underlying model architecture through the concept of a \textit{predictor}, supporting both single models and model \emph{ensembles} interchangeably. For ensembles, MUSE applies an intermediate \textit{Posterior Correction}~\cite{dalpozzolo2015calibrating} to mitigate the bias introduced by undersampling in imbalanced datasets. To ensure score distribution stability, MUSE enforces distributional invariance via \textit{Quantile Mapping}~\cite{panofsky1968some, bolstad2003comparison}. This latter score transformation maps model outputs to a pre-defined reference distribution without requiring labels. Finally, MUSE uses intent-routing to dynamically select the desired predictor and transformation sequence.

Our contributions are as follows:
\begin{itemize}
    \item A novel systems integration for stable model serving,
    combining Posterior Correction~\cite{dalpozzolo2015calibrating} to mitigate sampling bias and Quantile Mapping~\cite{panofsky1968some, bolstad2003comparison} to stabilize score distributions across model updates. This enables backward-compatible model evolution without access to client thresholds or labeled data.
    \item  We propose an intent-driven model serving abstraction designed specifically for Score-as-a-Service multi-tenancy. By treating routing, shadowing, and transformation updates as first-class deployment primitives, this design enables safe, automated, zero-friction model promotions in multi-tenant, real-time environments.
    \item We report results and operational insights from deploying MUSE in a real-world fraud detection application, in which thousands of events per second are processed with strict latency and availability requirements.
\end{itemize}

MUSE is deployed in production at Feedzai, currently serving several dozen of clients with a roadmap to scale to hundreds in a few months. In less than a year, MUSE has processed over 55 billion events across several use cases. Specifically in fraud detection, MUSE analyzed over \$1.8 billion in volume across 7 million transactions, preventing more than \$1.7 million in confirmed fraud\footnote{Figures based on two clients that provided feedback labels. This is the total confirmed fraud prevented in these clients, not the part attributable to faster model updates.}.

\section{MUSE}

MUSE is a distributed, low-latency, highly-available and high-throughput model serving framework designed to decouple the lifecycle of ML models from the downstream decision systems that consume them.
MUSE operates under strict availability and latency SLOs\footnote{MUSE \emph{Service-Level-Objectives} include a target latency lower than 30ms at p99 and 150ms at p99.9; and 99.95\% of availability.} while serving dozens of independent financial institutions (tenants) and processing thousands of events per second.
To achieve this at scale, MUSE adheres to four core design principles:

\begin{itemize}
    \item \textbf{Stateless Design:} All routing, orchestration, and transformation logic is stateless, allowing horizontal scaling via Kubernetes without state synchronization.
    \item \textbf{Multi-Tenancy \& Reuse:} Infrastructure is shared across predictors and model containers, optimizing costs.
    \item \textbf{Intent-Based Decoupling:} Clients request business intents (e.g., \emph{card fraud}) rather than model versions.
    \item \textbf{Composable Transformations:} Raw scores are not treated as static outputs but as malleable data. A configurable transformation pipeline adapts raw model outputs to stable business distributions.
\end{itemize}

\subsection{System Overview}
\label{sec:system_overview}

Figure~\ref{fig:diagram_1} presents the architecture of MUSE, a \textit{Directed Acyclic Graph} (DAG) pipeline. At its boundary, the system ingests raw transaction data and outputs a single fraud score. Features are computed internally through the Feature Store. We stress that information regarding the tenants' downstream decision system (and therefore their score thresholds) resides exclusively on the tenants' infrastructure and is not available to Feedzai. Within the DAG, requests flow through nodes representing discrete computational steps, such as model inference, aggregation, and score transformations. While our primary use case is fraud detection, the inference nodes are architecture-agnostic, supporting standalone models and/or complex ensembles, and can execute standard machine learning tasks, including classification and regression.

To optimize latency and resource usage, we decouple execution based on complexity. Lightweight operations (orchestration and transformations) run within a stateless workload, while compute-intensive inference is delegated to separate \textit{Model Servers} (e.g., NVIDIA Triton~\shortcite{triton}) on dedicated Kubernetes pods. This separation allows us to scale lightweight operations independently from the inference layer, ensuring cost-efficient infrastructure.

Crucially, the graph incorporates specific transformations: $\mathcal{T}^C$ for Posterior  Correction \cite{dalpozzolo2015calibrating} and $\mathcal{T}^Q$ for Quantile Mapping~\cite{panofsky1968some, bolstad2003comparison}. These components exist within the \textit{predictor}, the fundamental abstraction used by our intent-based routing mechanism to dynamically select the execution pipeline.

\subsection{Predictor Abstraction}
\label{sec:predictor_abstraction}

The core unit of our system is the \emph{predictor} $p$. A predictor is an abstraction that encapsulates any model(s) of any kind (e.g., LightGBM, XGBoost, Deep Learning, etc.). It receives a feature vector $\mathbf{x}$ and produces a business-ready score $\hat{y}$, hiding all the complexity from who calls the model:
\begin{equation}
\hat{y} = p(\mathbf{x}) .
\end{equation}
This score strictly refers to the model's output risk prediction (e.g., a probability between 0 and 1 for classification tasks). A predictor encapsulates a scoring DAG and hides the internal topology, whether it relies on a single model or a complex ensemble of heterogeneous learners.

\subsubsection{Efficient Ensemble Serving}
As detailed in Section~\ref{sec:model-serving-frameworks}, traditional model serving frameworks ~\cite{kserve, seldoncore, sagemaker} optimize for resource isolation over reuse, when serving transformations with ensembles. This causes infrastructure duplication, especially in multi-tenant settings, which MUSE solves by serving multiple inference DAG in the same workload.

As an illustrative example, consider Figure~\ref{fig:diagram_1}, in which predictor $p_1$ is composed of models $\{m_1, m_2\}$. To address a new fraud pattern, we deploy a new predictor $p_2$ that adds a specialized model $m_3$, resulting in the set $\{m_1, m_2, m_3\}$. Because $p_1$ and $p_2$ share the definitions for $m_1$ and $m_2$, MUSE reuses the existing containers for these models. The deployment of $p_2$ requires provisioning resources \textit{only} for the new model $m_3$, rather than a full replica of all three models.

This architecture provides two expected benefits:
\begin{itemize}
    \item \textbf{Infrastructure Deduplication:} Incremental updates to ensembles (adding or removing members) incur a marginal resource cost equal only to the net difference in models, rather than the total cost of the new ensemble.
    \item \textbf{Multi-Tenant Cost Savings:} A single model deployment can be referenced by hundreds of predictors. These predictors may be shared or they may contain client-specific transformations tailored to their data distribution, sharing the heavy computational cost of the underlying model.

\end{itemize}

\subsubsection{Ensemble Configuration}
MUSE supports ensembles by combining individual \emph{expert} models.
Let $\Gamma = \big\{\big(m_1, \mathcal{T}^C_1\big), \ldots, \big(m_K, \mathcal{T}^C_K\big)\big\}$ denote the set of $K$ expert models $m_k$ and their corresponding $\mathcal{T}^C_k$ posterior corrections (detailed in Section~\ref{sec:postcorrection}).
Formally, the predictor is defined as a tuple $p = \langle \mathcal{M}, \mathcal{A}, \mathcal{T}^Q \rangle$, where:
\begin{itemize}
    \item $\mathcal{M} \subseteq \Gamma$ is the subset of expert models consulted by this predictor (possibly just one);
    \item $\mathcal{A}$ is the aggregation function that combines the outputs of the selected models into a single score;
    \item $\mathcal{T}^Q$ is the quantile mapping that transforms the aggregated score such that it follows a stable, reference distribution (detailed in Section~\ref{sec:quantilemapping}).
\end{itemize}
Given an input feature vector $\mathbf{x}$, the final score $\hat{y}$ is given by:
\begin{equation}
\hat{y} = p(\mathbf{x}) =
\mathcal{T}^Q \bigg(
 \mathcal{A} \Big( \big[
  \mathcal{T}^C_k \big( m_k( \mathbf{x} ) \big)
 \big]_{(m_k, \mathcal{T}^C_k) \in \mathcal{M}} \Big) \bigg) \ .
\end{equation}

For single-model predictors ($\mathcal{M}=\{m\}$), the posterior correction is skipped and the aggregation function $\mathcal{A}$ is the identity, so the equation simplifies to $p(\mathbf{x}) = \mathcal{T}^Q \big( m(\mathbf{x}) \big)$.

\subsection{Composable Transformations}
\label{sec:transformations}

Within a predictor DAG, MUSE supports three possible score transformations: score calibration through Posterior Correction $\mathcal{T}^C$; ensemble aggregation $\mathcal{A}$; and score mapping through Quantile Mapping $\mathcal{T}^Q$, as illustrated in Figure~\ref{fig:diagram_1}.
Score calibration $\mathcal{T}^C$ and aggregation  $\mathcal{A}$ are typical of ensemble predictors and bypassed for single-models. The score mapping $\mathcal{T}^Q$ is applied to all predictors to ensure stable downstream behavior.

\subsubsection{Posterior Correction}
\label{sec:postcorrection}

When aggregating model scores, heterogeneous calibration distortions across individual models can act as implicit weights, biasing the ensemble aggregation~\cite{bella2013effect}. For instance, a model that systematically makes predictions closer to the extremes (0 and 1) will exert a disproportionate influence over the combined score, regardless of its predictive quality. Therefore, ideally all models are well-calibrated prior to aggregation~\cite{bella2013effect,kumar2022calibrated, fan2021applying}.

In domains such as fraud detection, in which datasets are heavily imbalanced, it is a common practice to undersample the majority (negative) class during model training in order to improve the learning efficiency and predictive performance.
However, the undersampling ratio affects the score distribution of the trained model: the more aggressively the majority class is undersampled, the higher the resulting scores tend to be when compared to versions trained on the original distribution.
To mitigate this sampling bias, MUSE applies a \emph{Posterior Correction} transformation~\cite{dalpozzolo2015calibrating} to the output of each expert model before aggregation, which rescales each model’s posterior probabilities.

Let $\tilde{y}_k$ be the raw score produced by expert model $m_k$, and let $\beta_k$ be the undersampling ratio of the majority negative class used during its training.
The calibrated score is:
\begin{equation}\label{eq:posterior_correction}
    \mathcal{T}^C_k(\tilde{y}_k) =
    \frac{\beta_k \tilde{y}_k}{1 - (1 - \beta_k) \tilde{y}_k} \ .
\end{equation}

By applying $\mathcal{T}^C_k$ independently to each expert’s output, prior to aggregation, MUSE removes the bias induced by the undersampling. Since this transformation is purely analytical, it introduces negligible latency overhead.

We note that Posterior Correction only removes the undersampling bias, and not calibration distortions resulting from the specific model choices. Alternative calibration methods (such as Platt Scaling~\shortcite{platt1999probabilistic} or Isotonic Regression~\shortcite{zadrozny2002transforming}) could be used in this step, but typically require a large amount of labeled data. This would incur other trade-offs, such as a reduced volume of training data and less recent training data.

\subsubsection{Ensemble Aggregation}

After computing the calibrated scores for the expert models, in case of an ensemble, these scores are combined to produce a single prediction for the client.
This is achieved through a configurable score aggregation step, which is part of the predictor's computation graph.

A common aggregation strategy is a weighted average of the calibrated scores.
The aggregation weights can be tuned for a client or use case, or default weights can be shared across multiple predictors.
Since these are simple arithmetic operations, the computational overhead is negligible.

Furthermore, tuning aggregation weights provides a lightweight mechanism to adapt predictors to new clients or shifting data distributions without retraining or redeploying underlying experts.
When labeled data becomes available, MUSE rapidly optimizes ensemble behavior at minimal cost while preserving expert reuse and shared infrastructure.

\subsubsection{Quantile Mapping}
\label{sec:quantilemapping}

MUSE applies quantile transformation as the last score transformation, before returning it to the client. It can be applied both for single-model and ensemble predictors, and it decouples the downstream business logic from the inference DAG that generated the score.

In practice, client systems convert risk scores into actions using fixed, user-defined thresholds.
Without additional constraints, even minor changes to the underlying models, aggregation logic, or training data can shift score distributions.
To avoid this, MUSE is designed so that the final output score aligns with a predefined reference distribution, independently of the predictor's internal structure.

For that, we use a \emph{Quantile Mapping}~\shortcite{panofsky1968some, bolstad2003comparison} transformation to align the Cumulative Distribution Function (CDF) of the predictor’s output score distribution $\mathcal{S}$ with that of a fixed reference distribution $\mathcal{R}$.
For computational efficiency, the mapping is approximated using a piecewise linear function defined over $N$ precomputed quantiles.

Let $q_1^\mathcal{S},\ldots,q_N^\mathcal{S}$ and $q_1^\mathcal{R},\ldots,q_N^\mathcal{R}$ denote the quantiles of the source and reference distributions, respectively.
Let $\tilde{y}$ be the score produced by the expert in a single-model predictor, or by the aggregation step in an ensemble predictor.
To map $\mathcal{S}$ to $\mathcal{R}$, for a given $\tilde{y}$ we find $i$ such that $q_i^\mathcal{S} \le \tilde{y} < q_{i+1}^\mathcal{S}$, which can be computed in $O(\log N)$ via binary search.
The mapped score is given by:
\begin{equation}\label{eq:quantile_transformation_linear}
    \mathcal{T}^Q(\tilde{y}) = q_i^\mathcal{R} + (\tilde{y} - q_i^\mathcal{S}) \cdot \frac{q_{i+1}^\mathcal{R} - q_i^\mathcal{R}}{q_{i+1}^\mathcal{S} - q_i^\mathcal{S}} \ .
\end{equation}

The resulting transformation is monotonic, so the ranking of the events is preserved (sorted by score), and, as such, the ranking-based predictive performance of the solution is not affected (Section~\ref{sec:discussion}).
The choice of reference distribution, $\mathcal{R}$, is fully configurable.
For example, in highly imbalanced settings such as fraud detection, choosing a reference distribution with a high density near $0$ and a longer tail towards $1$ allows the client to have more granularity on more useful regions of alert rates (typically between $0.1\%$ and $1\%$).
Alternatively, $\mathcal{R}$ can be chosen to match the score distribution of an existing production system, enabling the migration from legacy deployments.

Although the reference distribution is typically shared across predictors, the quantile mapping itself is tenant-specific.
Different clients exhibit different data distributions, and, as a result, predictors produce different source score distributions per tenant.
Consequently, the source quantiles $q_i^\mathcal{S}$ must be estimated separately for each client–predictor pair.

Accurate estimation of these quantiles requires a sufficient volume of unlabeled data.
As shown in Appendix~\ref{sec:bins_size}, for a target alert rate $a$, a maximum relative error $\delta$, and a confidence level corresponding to a z-score $z$ (e.g., $z = 1.96$ for $95\%$ confidence), the approximate number of samples $n$ is:
\begin{equation}
    n \approx \frac{z^2 (1 - a)}{\delta^2 a} \ .
    \label{eq:sample_size}
\end{equation}

This constraint dictates our cold-start strategy: for low-volume clients or new deployments, we use a default Beta-mixture initialization as a prior until sufficient live data exists.

\subsection{Cold-Start Problem}
\label{sec:zeroday}

We now address the cold-start problem, where we build a predictor capable of serving a new client without historical data, while still producing scores with a desired reference distribution $\mathcal R$. For this, we define a \emph{default} quantile transformation $\mathcal{T}_{v_{0}}^{Q}$. Since no data is available to describe the score distribution $\mathcal S$, which is needed to derive $\mathcal{T}_{v_{0}}^{Q}$, we replace $\mathcal S$ with a smooth  probability density function (PDF) $f_\mathcal S$.

To derive $f_\mathcal{S}$, we first compute the empirical distribution $f^{\text{emp}}_{\mathcal{S}}$, consisting of the predictor's score distribution on the combined training data of its expert models. Then, we fit a bimodal Beta mixture model to $f^{\text{emp}}_{\mathcal{S}}$. This choice is motivated by two factors: first, the Beta distribution provides natural support for the bounded $[0, 1]$ score interval; second, the mixture approach effectively models the bimodal density that typically characterizes scores in fraudulent and legitimate instances, ensuring smooth quantile estimation even in score regions with sparse training data.

Let $\tilde{y}$ denote the positive class score. We fit the PDF $f_{\mathcal{S}}$ to the data distribution $f^{\text{emp}}_{\mathcal{S}}$ as follows:
\begin{equation}\label{eq:bimodal_beta}
\begin{aligned}
    f_{\mathcal{S}}(\tilde{y}; \alpha_0, \beta_0, \alpha_1, \beta_1) &= (1 - w) \text{Beta}(\tilde{y}; \alpha_0, \beta_0) \\
    &\quad + w \text{Beta}(\tilde{y}; \alpha_1, \beta_1) .
\end{aligned}
\end{equation}
where $w = P(y=1)$ represents the prior probability of fraud in the dataset. The component $\text{Beta}(\tilde{y}; \alpha_0, \beta_0)$ approximates the class-conditional density $P(\tilde{y} \mid y=0)$, and $\text{Beta}(\tilde{y}; \alpha_1, \beta_1)$ approximates $P(\tilde{y} \mid y=1)$.

The shape parameters $(\alpha_i, \beta_i)$ are estimated by minimizing a cost function $\mathcal{L}$ representing the discrepancy between the $r$-th raw moments of the mixture model, $\mu_r(\alpha_0, \beta_0, \alpha_1, \beta_1)$, and the empirical moments of the $N$ training scores, $\bar{y}_r = \frac{1}{N} \sum_{i=1}^N \tilde{y}_i^r$:
\begin{equation}
    \mathcal{L} = \sum_{r=1}^4 \sqrt[r]{\left( \mu_r(\alpha_0, \beta_0, \alpha_1, \beta_1) - \bar{y}_r \right)^2} \, .
\end{equation}
The $r$-th root in the loss function evens out the importance of the moments, at the loss of differentiability, motivating the use of a stochastic search algorithm \cite{Storn1997}.
To reduce local minima sensitivity, the search is repeated across $N_\text{trial}$ runs, and the fit that minimizes the \textit{Jensen-Shannon Divergence} $\mathrm{JSD}$~\cite{lin2002divergence} against $f^{\text{emp}}_{\mathcal{S}}$ is selected:
\begin{equation}\label{eq:params_argmin}
\begin{aligned}
    (\alpha_0^*, \beta_0^*, \alpha_1^*, \beta_1^*) &= \operatorname{arg\,min} \Big[ \mathrm{JSD} \big( f^{\text{emp}}_{\mathcal{S}} \parallel \\
    &\quad f_{\mathcal{S}}(\tilde{y}; \alpha_0, \beta_0, \alpha_1, \beta_1) \big) \Big] .
\end{aligned}
\end{equation}

With the optimal parameters derived from Eqs.~\eqref{eq:bimodal_beta}-\eqref{eq:params_argmin}, the prior $f_{\mathcal{S}}$ lets us define a default quantile transformation $\mathcal{T}_{v_{0}}^{Q}$  when no data is available for $\mathcal S$.


\subsection{Routing}
\label{sec:routing}
Traditionally, Model Serving frameworks~\cite{kserve, seldoncore} couple clients to the model used for the prediction, by exposing the model as a request parameter~\cite{kserve_oip_2020}. As a result, model updates require client changes, demanding synchronized deployments across systems or organizations. In high-stakes environments, this requirement introduces operational risk: misaligned deployments cause service disruptions, and the coordination overhead delays model updates.

\subsubsection{Intent-Based Routing.}

To address this issue, MUSE leverages standard routing mechanisms common to service mesh architectures and API gateways~\cite{istio_docs_2026, traefik_docs_2026}. However, rather than routing based on model specific endpoints or headers, the system implements server side \textit{intent-based routing}: clients express their scoring intent (e.g., tenant id, geography, payment channel, etc.), and MUSE maps it to the appropriate predictor. Listing~\ref{lst:routing_yaml} shows an example of routing configuration in MUSE, where teams define \textit{scoring rules} -- evaluated sequentially to identify which predictor is used to serve which intent; and \textit{shadow rules} -- evaluated in parallel (multiple shadow rules may trigger) to determine which predictors requests are mirrored to, so their responses can be stored in a Data Lake, without affecting the returned response. For example, in Listing~\ref{lst:routing_yaml}, a request from \texttt{bank1} is served by \texttt{bank1-model-v1}, while being sent to \texttt{bank1-model-v2} for offline evaluation purposes.

\begin{listing}[t]
\caption{Example of declarative routing configuration.}
\label{lst:routing_yaml}
\begin{lstlisting}[language=yaml,numbers=none,xleftmargin=0pt,columns=fullflexible,keepspaces=true]
routing:
  scoringRules:
  - condition: {tenants: ["bank1"]}
    targetPredictorName: "bank1-model-v1"
  - condition:
      geographies: ["NAMER", "LATAM"]
      schemas: ["fraud_v1"]
    targetPredictorName: "us-predictor-v1"
  - condition: {}  # Default catch-all
    targetPredictorName: "global-v3"
  shadowRules:
  - condition: {tenants: ["bank1"]}
    targetPredictorNames: ["bank1-model-v2"]
\end{lstlisting}
\end{listing}

Because the routing logic relies exclusively on request metadata without requiring external lookups or state synchronization, it introduces negligible overhead. Once the appropriate predictor is selected, the system may enrich the request by querying a feature store for any additional model-specific features not included in the initial payload. Furthermore, because the \textit{predictor} abstraction can represent either a single model or an ensemble of models, and optionally contain a set of transformations (e.g., Posterior Correction and Quantile Mapping), routing rules can map to any part of the inference pipeline, depending on what best fits the business needs. In this way, MUSE supports client-free intervention for the following model lifecycle operations:

\begin{enumerate}
    \item \textbf{Transparent Model Switching:} Any modification to a predictor, i.e., changing a model, a transformation, or the weights of an ensemble, is executed via a single update to a server-side routing rule. Since the client continues to request the same intent, the transition is transparent from the client's perspective, requiring no client-side code changes or maintenance windows.
    \item \textbf{Shadow Scoring:} Routing supports a 1-to-N resolution, where a request is resolved to exactly one \textit{Live} predictor (for the client response) and an arbitrary number of \textit{Shadow} predictors. This allows us to validate multiple predictor versions simultaneously on the same production stream without interfering with live traffic.
    \item \textbf{Easy Feature Evolution:} Models with heterogeneous feature sets can be supported simultaneously. E.g., one can test or promote new model versions without client intervention. MUSE automatically identifies the new model's derived features and ensures that it receives the proper enriched feature set from the feature store.
    \item \textbf{Scalable Multi-Tenancy:} Managing multiple transformation pipelines, model versions, and ensembles for hundreds of tenants introduces significant complexity. MUSE abstracts this by allowing routing rules to be either highly granular (mapping tenant segments to unique predictors) or generalized (grouping similar tenants to the same transformations). This centralized control contrasts with decentralized approaches where configuration is scattered across client integrations, allowing MUSE to provide a comprehensive view of all tenants and policies, alongside their associated predictors and transformation pipelines.
\end{enumerate}

\subsubsection{Rolling Deployments and Consistency.}
Because the routing layer is stateless, model promotions execute via standard Kubernetes Rolling Updates.
Upon routing changes (e.g., promoting a shadow model to live), the MUSE control plane detects the update and initiates a rolling restart.

\begin{figure}
\centering
    \includegraphics[width=0.9\columnwidth]{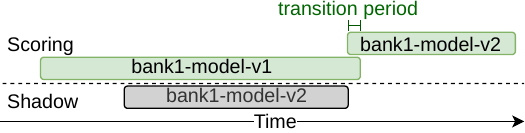}
    \caption{The model lifecycle: from training to shadow validation, and finally to live promotion.}
    \label{fig:model_lifecycle}
\end{figure}

Figure \ref{fig:model_lifecycle} illustrates the lifecycle for \texttt{bank1}, from the previous configuration: \texttt{bank1-model-v2} is first deployed in shadow mode to validate latency and distribution stability; after an evaluation period, it is promoted to live scoring, and \texttt{bank1-model-v1} is decommissioned. A predictor is only promoted after its shadow scores confirm the expected distribution. Since configurations are versioned in Git, rollback to a previous configuration amounts to reverting a commit.

\section{Evaluation}

We evaluate MUSE on a multi-tenant production environment serving dozens of financial institutions globally. It operates under strict SLOs of 30\,ms p99 latency and 99.95\% availability, and processes an average of 4{,}500 events per second in the EU region alone, with a measured availability of 99.99\% over the last 30 days. Although MUSE is deployed in additional regions and use cases, this evaluation focuses on Transaction Fraud across a subset of 4 clients with varying characteristics on 3 different use cases, as detailed in Table~\ref{tab:client_metrics}.

\begin{table}[t]
    \centering
    \small
    \setlength{\tabcolsep}{3pt}
    \begin{tabular}{lccccc}
        \toprule
        \textbf{Client} & \textbf{A}  & \textbf{B} & \textbf{A} & \textbf{C} & \textbf{D} \\
        \midrule
        \textbf{Use Case} & 1 & 2 & 3 & 3 & 3 \\
        \textbf{\makecell[l]{Avg. Transactions\\per Day}} & 231760 & 256434 & 33568 & 2712112 & 616 \\
        \textbf{Weeks of Data} & 10 & 10 & 10 & 10 & 5 \\
        \textbf{Ensemble Size} & 5 & 8 & 3 & 3 & 4 \\
        \textbf{Evaluation Windows} & 5 & 5 & 5 & 5 & 2 \\
        \bottomrule
    \end{tabular}
    \caption{Deployment metrics across anonymized clients.}
    \label{tab:client_metrics}
\end{table}

\subsection{Live Predictor Updates}

We analyze the effectiveness of quantile mapping in two common production scenarios: 1) a transformation update, by transitioning from a default transformation to a client-specific one; 2) a model update, by adding a new model to an ensemble. For each scenario, we fit a new $\mathcal{T}^Q$ on 14 days preceding each deployment and we evaluate the resulting calibration for the following 7 days. We perform this analysis over a 3 month period for each entry in Table~\ref{tab:client_metrics}, yielding 22 client--update-point experiments, and report the relative error per score bin against the target distribution\footnote{The target distribution used in MUSE was developed based on internal expertise and is proprietary company information.}, pooled over all experiments. We complement this analysis with a public, reproducible version of the same experiment, run on the open Bank Account Fraud dataset~\cite{jesus2022turning} and publicly available on GitHub\footnote{\url{https://github.com/feedzai/muse-quantile-study}}, where, due to label scarcity, we also show a precision and recall comparison against a label-based recalibration baseline.

For each scenario and each client, every predictor variant is scored on the same 7-day evaluation window, so predictors are always compared on identical events. Scores are grouped into ten disjoint bins $b$; the relative error of bin $b$ is $(\hat s_b - s_b)/s_b$, where $\hat s_b$ is the observed share of events in $b$ and $s_b$ the share prescribed by the reference distribution $\mathcal{R}$. Boxes show median and interquartile range (IQR), and whiskers the 5th--95th percentiles. Our goal is to keep each client's operating point stable when the model changes, and for this purpose we make use of two baselines: serving the raw model, and updating a model without retuning client thresholds.

\subsubsection{Default to Client-Specific Transformation}
\label{sec:eval:zeroday}

\begin{figure}
    \centering
    \includegraphics[width=\columnwidth]{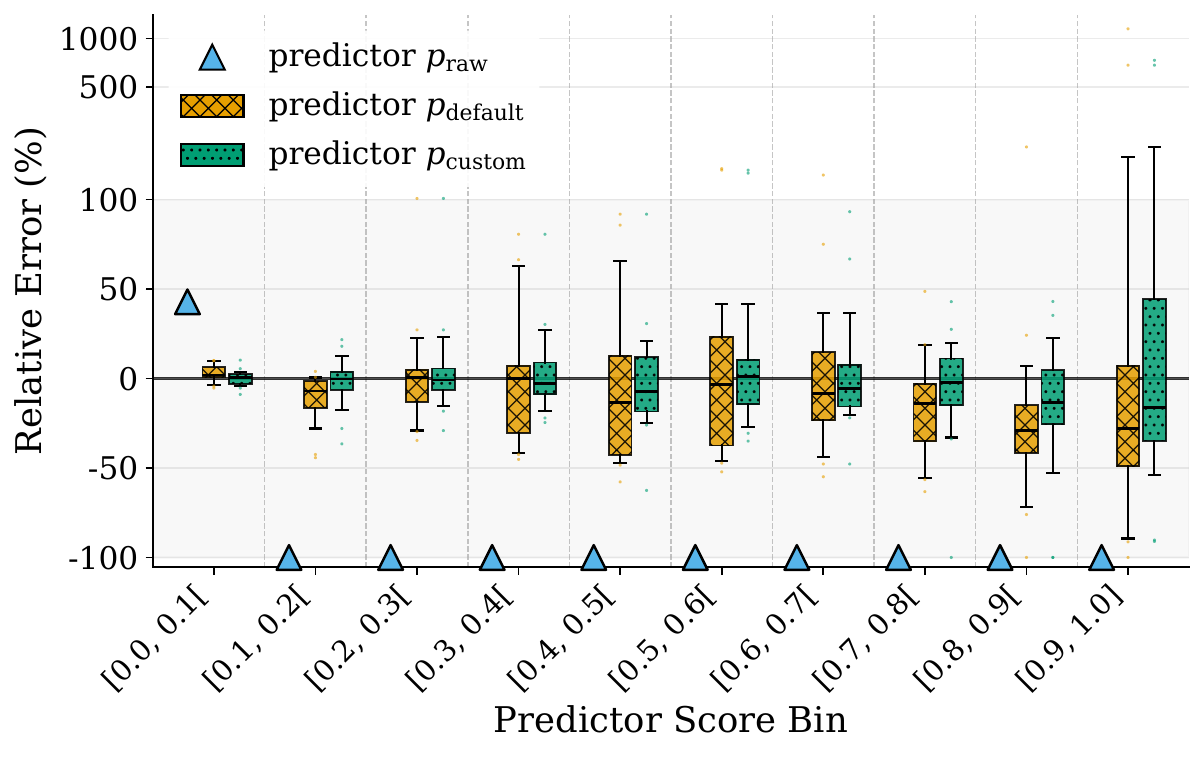}
    \caption{Transformation Update - relative error to target alert rate when replacing the default with a custom transformation.}
    \label{fig:transformation_update_error}
\end{figure}

MUSE typically applies a default transformation $\mathcal{T}_{v_{0}}^{Q}$ during the onboarding period of new clients, so it can deliver immediate value by scoring transactions from day one without requiring historical data. For example, Client B was served $\mathcal{T}_{v_{0}}^{Q}$ during the first 2 weeks of its onboarding period, scoring over 2 million transactions, totaling \$165.7 million. After 2 weeks, with sufficient live data, the system began to periodically refit and apply client specific transformations $\mathcal{T}_{v_{t}}^{Q}$.

Figure~\ref{fig:transformation_update_error} compares the relative error of the observed alert rate per score bin, for all Table~\ref{tab:client_metrics} clients.
For that, we use three predictors that share the same expert models $\mathcal{M}$, but with different quantile transformations: $p_{\text{default}}$ using the default transformation $\mathcal{T}_{v_{0}}^{Q}$; $p_{\text{custom}}$ with a client-tuned transformation $\mathcal{T}_{v_{t}}^{Q}$; and finally $p_{\text{raw}}$ without any $\mathcal{T}^Q$.

As observed, $p_{\text{raw}}$ scores exhibit extreme skew, concentrating nearly all volume below the 0.1 threshold. Consequently, for the remaining thresholds, the observed alert rate plummets to near zero, registering a $-100\%$ relative error. This extreme deviation shows that raw outputs cannot be reliably used in decision rules, as they would often cause over or under alerting. In contrast, $p_{\text{default}}$ is able to produce relatively stable alert rates, with median errors within $\pm 15\%$ up to $0.8$ bin and of about $-29\%$ in the two top bins. As expected, $p_{\text{custom}}$ exhibits even smaller errors, maintaining a median error below 17\% across all thresholds, including the top bins. By refitting the quantile mapping using the most recent live traffic, $p_{\text{custom}}$ restores and maintains a tighter alignment with the target alert rate. The extreme points in the top bins mostly come fro Client D, which averages $\approx$4.3k events per evaluation week: it is fitted on the same 14-day window, below the volume Eq.~(\ref{eq:sample_size}) requires for the lowest alert rates, and its upper bins expect only a handful of events per week. Appendix~\ref{sec:extra_results} breaks down the $p_{\text{custom}}$ results by client.

\subsubsection{Ensemble and Transformation Update}
\label{sec:live_model_update}

\begin{figure}
    \centering
    \includegraphics[width=\columnwidth]{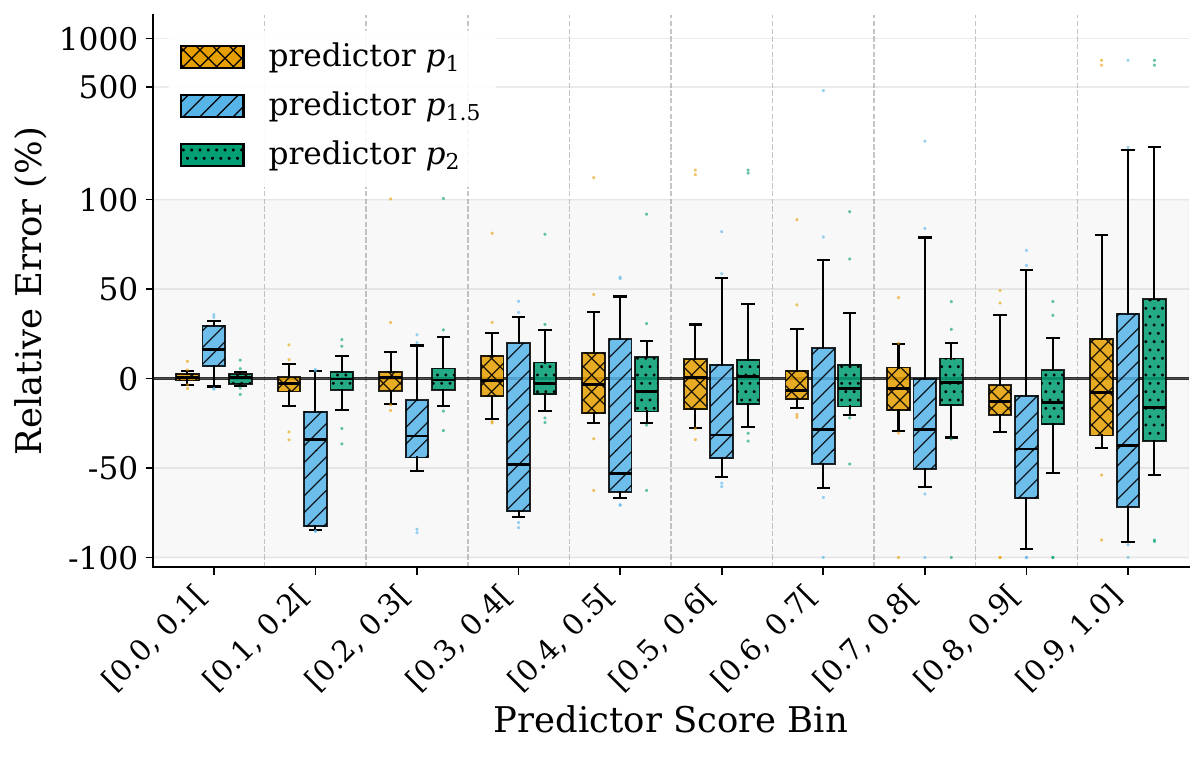}
    \caption{Expert Update - relative error to target alert rate. Updating an expert without refreshing the transformation ($p_{1.5}$) causes under-alerting, whereas pairing it with an updated transformation ($p_2$) restores alignment.}
    \label{fig:model_update_drift}
\end{figure}

In a second scenario, we evaluate the score distribution shift caused by adding a new expert model to an ensemble $\mathcal{M}$, producing a new ensemble $\mathcal{M}'$. This affects the underlying raw score distribution for all clients utilizing $\mathcal{M}'$, and, to maintain alert rates close to the target mapping, MUSE must refit  each quantile transformation individually for each client.

Figure~\ref{fig:model_update_drift} compares the relative error rate across three predictors: \emph{predictor $p_1$}, using $\mathcal{M}$ and its client-specific transformation; \emph{predictor $p_{1.5}$}, a hypothetical scenario where  $\mathcal{M}'$ is deployed but the original $\mathcal{M}$  transformation is kept; and \emph{predictor $p_2$}, where $\mathcal{M}'$ is paired with a new transformation fit on that ensemble's raw distribution.

As expected, \emph{predictor $p_{1.5}$} exhibits severe misalignment, reaching an Interquartile Range (IQR) of up to 108, with Q1 below $-44\%$ in every bin above $0.1$. Adding an expert without updating the transformation causes significant under-alerting, particularly for thresholds above 0.7, potentially reducing fraud coverage and requiring client-side threshold retuning. In contrast, both \emph{predictors $p_1$ and $p_2$} maintain close alignment with the target distribution, with relative errors mostly close to 0\% and a worst-case median of only $-16.1\%$. This demonstrates that MUSE can update the underlying served model without observable impact on client alert rates, allowing improved models to reach production sooner. For instance, in the scenario above, the updated ensemble $\mathcal{M}'$ yielded a 1.1 percentage point increase in Recall at a fixed 1\% False Positive Rate (FPR). Based on historical deployment timelines, promoting a model to production typically takes weeks; for a particular MUSE client, deployment was accelerated by 10 weeks, effectively reducing the promotion step to minutes. Lead time here runs from a validated model to live scoring: MUSE still collects 14 days of scores to fit the client-specific mapping, which must pass a quality check, followed by a few days of shadow scoring. This reduction is achieved by bypassing manual operational steps: client synchronization, deployment scheduling, rollout monitoring, and iterative threshold tuning. Eliminating this overhead accelerates the deployment of more accurate models with higher recall gains, which has been shown to prevent millions of dollars more in fraud~\cite{BrancoAGAAB20}.

\subsubsection{Operational Stability}
In addition to score distribution stability, the system must ensure operational performance during updates. A primary challenge is service warm-up, because Java workloads typically suffer from high latencies during Just-In-Time (JIT) compilation periods, most common on fresh workloads. To avoid exposing clients to this degradation, we perform \emph{code warm-up} at startup, by ``simulating'' requests and forcing compiler optimizations.

Before a Kubernetes pod is set as \emph{ready} to receive requests, a subprocess is launched with the configuration of all predictors. This subprocess invokes remote calls with synthetic data to the main program, simulating real requests. After the desired number of warm-up calls, the subprocess exits and sends a signal to the main process indicating the service is ready. By then, the hot-path methods are already optimized, leading to improved latency and throughput performance.

All MUSE transformations and model updates are performed through rolling deployments to ensure high-availability. During this rollover period (shown in Figure~\ref{fig:model_lifecycle}), traffic is split between instances when a rollover occurs.

\begin{figure}[t]
    \centering
    \includegraphics[width=\columnwidth]{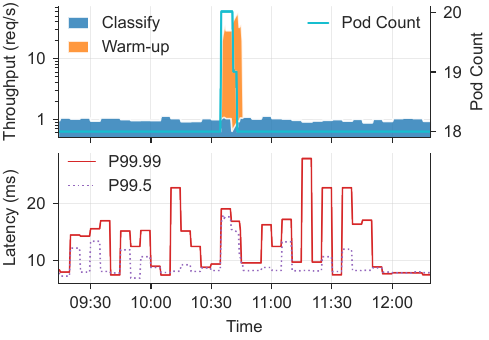}
    \caption{System performance during the update from $p_{\text{default}}$ to $p_{\text{custom}}$. During a rollover, the warm-up process ensures strict adherence to latency targets throughout the transition.}
    \label{fig:xp2_performance}
\end{figure}

Figure~\ref{fig:xp2_performance} details system latency and throughput during the experiment described in section~\ref{sec:eval:zeroday} for Client B, during the transition period between $p_{\text{default}}$ and $p_{\text{custom}}$. The top plot shows the pod count increasing and returning to baseline. The initialization of each new pod triggers the warm-up procedure, with a 15-minute warm-up procedure triggering spikes up to 50 req/s.
Without this warm-up procedure, benchmarks show that cold JVMs had p99.99 latencies of up to 1\,s during the same transition, an order of magnitude above the 30\,ms SLO. The bottom plot depicts system latencies across the p99.99 and p99.5 percentiles. It can be seen that latencies remained below 30ms for these percentiles throughout the update, demonstrating that MUSE can execute complex transformation swaps without latency degradation. This trace covers a single rollover for one client; but in the last last 30 days, MUSE measured across all EU traffic, over 5.0 billion requests, p99 latency of 6.7\,ms and a p99.9 of 84\,ms.

\subsection{Expert Calibration}
\label{sec:ensemble_normalization}

This section analyzes the effect of Posterior Correction on an ensemble of three models $\{m_1, m_2, m_3\}$, each with its respective undersampling rate $\beta$. We compute the calibration error in two scenarios: \textit{in-distribution} validation data and \textit{out-of-distribution} live client data. For each scenario, we present two calibration metrics: the Expected Calibration Error (ECE)~\cite{Pakdaman_Naeini_Cooper_Hauskrecht_2015, pavlovic2025understanding, guo2017calibration, pmlr-v151-roelofs22a} and the Brier score~\cite{glenn1950verification, hernandez2012unified}. ECE measures a model's calibration by comparing its predicted probabilities to the prevalence of the positive class at such probabilities. In particular, we use the $\text{ECE}^\text{EM}_\text{SWEEP}$ estimator\footnote{EM stands for Equal Mass, and SWEEP for finding the optimal number of bins.}~\cite{pmlr-v151-roelofs22a}, as it is less biased than the standard ECE estimator. However, since a model with constant predictions can trivially achieve a perfect $\text{ECE}=0$~\cite{pavlovic2025understanding}, we complement with the Brier score (or Mean Squared Error), measuring both calibration and predictive performance.

\begin{table}[t]
    \centering
    \small
    \setlength{\tabcolsep}{3pt}
    \renewcommand{\arraystretch}{1.2}
    \begin{tabular}{cl|lccc} \noalign{\hrule height 0.8pt}
        & \textbf{Model ($\beta$)} & \textbf{Error} & \textbf{w/o PC} & \textbf{w/ PC} & \textbf{$\Delta$ (\%)} \\
        \noalign{\hrule height 0.8pt}

        \multirow{6}{*}{\rotatebox[origin=c]{90}{\textbf{Validation}}}
        & \multirow{2}{*}{$m_1$ ($\approx 18\%$)}
        & ECE   & \num{4.97e-3} & \textbf{\num{7.67e-4}} & $\mathbf{-84.6}$ \\
        & & Brier & \num{1.46e-4} & \textbf{\num{9.61e-5}} & $\mathbf{-34.2}$ \\
        \cline{2-6}

        & \multirow{2}{*}{$m_2$ ($\approx 18\%$)}
        & ECE   & \num{4.83e-2} & \textbf{\num{8.63e-3}} & $\mathbf{-82.2}$ \\
        & & Brier & \num{2.70e-3} & \textbf{\num{4.36e-4}} & $\mathbf{-83.9}$ \\
        \cline{2-6}

        & \multirow{2}{*}{$m_3$ ($\approx 2\%$)}
        & ECE   & \num{4.23e-2} & \textbf{\num{8.46e-4}} & $\mathbf{-98.0}$ \\
        & & Brier & \num{1.86e-3} & \textbf{\num{1.58e-5}} & $\mathbf{-99.1}$ \\
        \hline
        \hline

        \multirow{8}{*}{\rotatebox[origin=c]{90}{\textbf{Live Client}}}
        & \multirow{2}{*}{$m_1$ ($\approx 18\%$)}
        & ECE   & \num{1.35e-2} & \textbf{\num{2.29e-3}} & $\mathbf{-83.0}$ \\
        & & Brier & \num{6.76e-4} & \textbf{\num{1.94e-4}} & $\mathbf{-71.4}$ \\
        \cline{2-6}

        & \multirow{2}{*}{$m_2$ ($\approx 18\%$)}
        & ECE   & \num{4.82e-2} & \textbf{\num{8.77e-3}} & $\mathbf{-81.8}$ \\
        & & Brier & \num{2.54e-3} & \textbf{\num{2.55e-4}} & $\mathbf{-89.9}$ \\
        \cline{2-6}

        & \multirow{2}{*}{$m_3$ ($\approx 2\%$)}
        & ECE   & \num{6.85e-2} & \textbf{\num{1.24e-3}} & $\mathbf{-98.2}$ \\
        & & Brier & \num{5.04e-3} & \textbf{\num{1.77e-4}} & $\mathbf{-96.5}$ \\
        \hhline{~=:====}

        & \multirow{2}{*}{\makecell{Ensemble\\$\{m_1,m_2,m_3\}$}} & ECE & \num{4.25e-2} & \textbf{\num{3.93e-3}} & $\mathbf{-90.8}$ \\
        & & Brier & \num{2.05e-3} & \textbf{\num{1.92e-4}} & $\mathbf{-90.6}$ \\
        \noalign{\hrule height 0.8pt}
    \end{tabular}
    \caption{Brier score and ECE calibration errors (lower is better) before (w/o PC) and after (w/ PC) applying Posterior Correction. Results are presented for each individual expert on their validation datasets, and for both the individual experts and the final ensemble on live client data. The $\Delta$ (\%) column denotes the relative percentage change in error, with negative values indicating a performance improvement.}
    \label{tab:pc-data}
\end{table}

Table~\ref{tab:pc-data} shows the calibration improvements from Posterior Correction. For individual experts on their respective validation data, the correction reduces errors significantly, with ECE decreasing by more than $80\%$ and Brier scores by more than $30\%$, experimentally corroborating.~\citet{dalpozzolo2015calibrating} on our real fraud detection scenario.

In the second scenario we operate outside standard Posterior Correction assumptions, as the used live client data is out-of-distribution. However, given the strong downsampling of up to two orders of magnitude (e.g. $\beta \approx 2\%$), the undersampling bias correction becomes a dominant calibration factor even out-of-distribution.
Effectively, the calibration improvements closely match the in-distribution results, especially for ECE calibration error which again decreases over $-80\%$ with very similar results for each expert, while Brier scores decrease over $-70\%$. The positive impact of Posterior Correction in the final ensemble is visible at the bottom of Table~\ref{tab:pc-data}. On live data, the ensemble of calibrated experts reduced both errors by over $-90\%$, even with $\beta$ varying by one order of magnitude between experts ($2\%$ vs. $18\%$).

\section{Discussion and Limitations}
\label{sec:discussion}

Quantile mapping matches the \emph{marginal} score distribution of each tenant to $\mathcal{R}$, assuming the fitting window is representative of the traffic that follows. We highlight what this does and does not guarantee.

\begin{itemize}
    \item \textbf{Ranking:} The mapping is monotonic, so ranking-based metrics (e.g., recall at a fixed FPR) are unchanged for a fixed model.
    \item \textbf{Alert rate:} A tenant's overall alert rate at any fixed threshold stays within the relative tolerance $\delta$ of Eq.~(\ref{eq:sample_size}) across model updates, provided the fitting window holds the number of events Eq.~(\ref{eq:sample_size}) requires and each update is paired with a refit. In our evaluation the median error stayed below 17\% in every bin (Section~\ref{sec:eval:zeroday}), while a tenant below that volume (Client D) deviated far more. This does not hold under arbitrary drift between refits, nor within segments that a tenant's rules combine with other features (e.g., amount), whose volumes may still change after an update.
    \item \textbf{Calibration and genuine risk changes:} The mapped score is not a calibrated fraud probability. Since mappings are only refit when the predictor changes, a genuine increase in fraud risk between refits still raises alert volumes, as intended; a refit on a window that contains an attack would instead absorb it.
    \item \textbf{Sample size:} Eq.~(\ref{eq:sample_size}) assumes independent draws from a stationary distribution and bounds only the fitting error (Appendix~\ref{sec:bins_size}).
    \item \textbf{Posterior Correction:} Posterior Correction removes the bias that undersampling introduces at training time, ensuring experts are calibrated before they are aggregated. While it reduced the ensemble's ECE and Brier score by over 90\% on live client data (Section~\ref{sec:ensemble_normalization}).    
\end{itemize}

\section{Related Work}
\label{sec:relatedwork}
Existing work addresses model serving, score calibration, and ensemble correction largely in isolation.
MUSE operates at their intersection, motivated by the need to perform continuous model updates while maintaining score stability.

\emph{Model Serving Frameworks.}\label{sec:model-serving-frameworks}
The Model Serving ecosystem can be split into specialized \textit{model servers} and higher-level \textit{serving frameworks}. Model servers, such as NVIDIA Triton~\shortcite{triton}, TensorFlow Serving~\shortcite{tfserving}, and MLServer~\shortcite{mlserver}, focus on high-performance inference but lack higher-level primitives. MUSE uses Triton~\shortcite{triton} as its inference engine, leveraging it to serve all predictors' models.

Serving frameworks, encompassing managed platforms like SageMaker~\shortcite{sagemaker}, Vertex AI~\shortcite{vertex_ai}, Azure ML~\shortcite{azure_ml}, and Snowflake~\shortcite{snowpark}, as well as open-source solutions such as KServe~\shortcite{kserve}, Seldon Core~\shortcite{seldoncore}, Ray Serve~\shortcite{rayserve}, MLflow~\shortcite{mlflow}, and BentoML~\shortcite{bentoml}, feature standard MLOps operations (e.g., model deployment, governance, monitoring, etc.). However, they lack support for complex calibration pipelines. Some open-source frameworks offer composable \emph{routing} between model, but do not treat \emph{transformations} as first-class, composable graph nodes in the same way. For example, KServe (v0.16) enforces a 1:1 mapping between each model and its transformer, rather than reusing several independent, tenant-specific transformations in its \emph{InferenceGraph}. As a result, serving the same ensemble to multiple clients with unique calibrations requires deploying a separate \emph{Inference Service} per tenant. With dozens of tenants sharing ensembles, this 1:N duplication means running roughly $2N$ deployments for $N$ tenants (accounting for standard high-availability replication), which would mean using hundreds of duplicated CPU allocations and dozens of additional cluster IPs consumed purely for per-tenant transformer instances. For reference, MUSE serves 201 tenants and 46 predictors with 23 pods for one of its use cases (some clients are also multi-tenant) , with the same shared model servers. A per-predictor transformer layout would require an estimated 108 pods (4.7$\times$) and 2.4$\times$ the CPU. These figures are an estimate derived from the deployed configuration.
MUSE avoids this duplication by decoupling transformations from the model runtime. It aggregates all client-specific transformations into a single deployment while sharing the underlying model servers.

\emph{Commercial Fraud Score Providers.}
Commercial solutions such as Stripe Radar~\shortcite{stripe_radar} and Kount~\shortcite{kount_omniscore} use \textit{probabilistic scoring} anchored to a \emph{global} fraud probability (e.g., a score of 90 implies a 90\% fraud likelihood).
This approach couples the tenant's decision volume directly to the global threat landscape; if a fraud attack spikes, high-risk transaction volume surges, violating operational capacity constraints.
Sift~\shortcite{sift_score_api} mitigates this via a secondary percentile score over rolling traffic windows alongside raw model output, increasing the delivered risk signal's complexity. MUSE avoids rolling score windows by enforcing distributional invariance against a \textit{fixed} reference, guaranteeing stable operations without shifting complexity to clients. We do not evaluate against Sift because by construction it always absorbs genuine increases in fraud risk, and MUSE keeps the distribution stable but is still capable over alerting when fraud activity increases.

\emph{Distributional Stability.}
Predictor updates often cause compatibility issues when distributions diverge, improving performance but breaking downstream systems~\shortcite{srivastava2020empirical}.
The most common approach to map any observed distribution onto a predetermined one is Quantile Mapping, with applications in meteorology~\shortcite{panofsky1968some}, digital image processing~\shortcite{gonzalez2018digital}, and genomics~\shortcite{bolstad2003comparison}.
Adopting this ensures scores match the same percentile, keeping client alert rates stable across model updates without changing business logic.

\emph{Expert Calibration in Ensembles.}
Training experts with different undersampling rates introduces distinct biases, distorting their influence within the ensemble.
This bias can be corrected with Posterior Correction~\shortcite{dalpozzolo2015calibrating}, a statistical technique grounded in earlier work~\cite{saerens2002adjusting, elkan2001foundations}.
Unlike learnable calibration methods (e.g., Platt Scaling~\shortcite{platt1999probabilistic} or Isotonic Regression~\shortcite{zadrozny2002transforming}), Posterior Correction requires only the training sampling rate and no labeled data, making it ideal for environments with label delay. While well-suited for our purpose, Posterior Correction only reverses the sampling bias without addressing any model-specific calibration distortions.

\section{Conclusions}

We present MUSE, a model serving framework to streamline the lifecycle of ML models in multi-tenant \textit{Score-as-a-Service} environments. MUSE treats model evolution as more than an infrastructure change. It prioritizes score stability to establish a reliable contract where model upgrades do not invalidate the downstream business rules that rely on them.
MUSE is deployed in production at Feedzai, serving several dozens of financial institutions. In the fraud detection use case evaluated in this paper, the system has processed over \$1.8 billion in transaction volume. In two clients that provide labels, MUSE-served models prevented over \$1.7 million in confirmed fraud losses in a few months of operation, while reducing time-to-production for more capable models without the delays inherent of client coordination. The system achieves this while adhering to strict high-availability and latency constraints, handling thousands of events per second across multiple use cases with negligible overhead from the transformation pipeline.

As we scale to hundreds of clients, we aim to evolve MUSE in two directions: 1) \textit{Automated Calibration Refresh}: currently, transformation updates are triggered by deployments. We plan to automate background Quantile Mapping re-fitting based on closed-loop distribution drift monitoring, ensuring stability between retrains. 2) \textit{Generalized Posterior Correction}: we aim to investigate generalized correction methods to dynamically balance experts more effectively, using the undersampling rate alongside other heuristics (e.g., training volume, validation performance, recency).

\section*{Acknowledgments}
We thank Nikoletta Matsur, Javier Liébana and Artyom Korkhov for their contributions on initial versions of this system.

\bibliography{bibliography}

\appendix\label{ap:uncertainty_estimation}

\section{Statistical Derivation of Sample Size Requirements for Quantile Transformation}
\label{sec:bins_size}

In this section, we statistically derive the minimum number of events $n$ required during the fitting period of the Quantile Transformation ($\mathcal{T}^Q$), described in Equation~\ref{eq:number_of_points}.
A major source of uncertainty in the alert rate of the Quantile Transformation is the stochastic nature of threshold selection.

\subsection{Fitting Phase}
During the fitting phase of $\mathcal{T}^Q$, we draw $n$ scores from an unknown distribution $f$. To achieve a target alert rate $a$, the $k$-th lowest score is selected as the decision threshold, such that the proportion of alerted scores (above the threshold) is approximately $\frac{n-k}{n} \approx a$.

By the probability integral transform, sampling a score $x$ from $f$ with CDF $F$ is equivalent to sampling $u \sim \text{Uniform}(0, 1)$ and computing $x = F^{-1}(u)$. The $k$-th lowest score corresponds to the $k$-th order statistic $U_{(k)}$ in the uniform space, which follows a $\text{Beta}(k, n-k+1)$ distribution with density:
\begin{equation}
\begin{aligned}
f_{U_{(k)}}(u) &= \frac{n!}{(k-1)!(n-k)!} u^{k-1}(1-u)^{n-k}, \\ \quad u &\in [0,1].
\end{aligned}
\end{equation}

Expectation and variance of the threshold are:
\begin{align}
\mathbb{E}[U = u_{(k)}] &= \frac{k}{n+1} \approx 1-a
\label{eq:mean_of_threshold}
\\
\text{Var}(U = u_{(k)}) &= \frac{k(n-k+1)}{(n+1)^2(n+2)} \approx \frac{a(1-a)}{n} \label{eq:variance_of_threshold}
\end{align}
by applying the asymptotic approximation for large $n$, and the approximation $\frac{n-k}{n} \approx a$.

Following the same approximation for large $n$, each sample quantile follows a normal distribution with mean and variance from Equations~\eqref{eq:mean_of_threshold}~and~\eqref{eq:variance_of_threshold}.
Thus, to determine the minimum sample size $n$ such that the observed alert rate will be within a certain relative deviation $\delta$ from the target alert rate $a$, we can use the formula for the confidence intervals of the normal distribution, where $z$ depends on the confidence level (e.g. for 95\% confidence, $z=1.96$).
\begin{align}
    (1-a) \pm \delta a &= \mu \pm z\sigma \\
    \Leftrightarrow (1-a) \pm \delta a &= (1-a) \pm z \sqrt{\frac{a(1-a)}{n}}
\end{align}

Solving for n, we get:
\begin{equation}
\delta a = z \sqrt{\frac{a(1-a)}{n}} \quad \Longrightarrow \quad n = \frac{z^2 (1-a)}{\delta^2 a}.
\label{eq:number_of_points}
\end{equation}

This equality establishes a lower bound on the fitting sample size $n$ required to guarantee a maximum relative error $\delta$ from the target alert rate $a$ with confidence level corresponding to $z$, assuming the $n$ scores are independent draws from a stationary distribution; it does not account for drift, nor for the sampling noise of the window on which alert rates are later observed.

We note that the normal approximation for sample quantiles generally requires $n a \gg 1$. However, substituting Equation~\eqref{eq:number_of_points} yields $na \approx z^2/\delta^2$. For standard confidence levels (e.g., $z=1.96$) and practical relative error bounds (e.g., $\delta \leq 0.2$), the normality condition is naturally satisfied ($na \approx 100$).

\section{Quantile Transformation Algorithm}

\subsection{Fitting the transformation}
Let $\mathcal{S} = \{s_1, s_2, \dots, s_n\}$ denote the set of raw scores collected over a period for a given client.

We define a set of source quantiles $[q_1, q_2, \dots, q_6]$, not necessarily uniformly spaced, that we want to map to a set of target scores $ [q^\mathcal{R}_1, q^\mathcal{R}_2, \dots, q^\mathcal{R}_6]$.

For each quantile ($q_k$), we compute the corresponding score $q^\mathcal{S}_k$ in the original score distribution $\mathcal{S}$:
\begin{equation}
q^\mathcal{S}_k = \text{Quantile}(\mathcal{S}, q_k) \label{quantile_map_eq}
\end{equation}

This yields a mapping from a set of original scores $\{q_k\}$ to a set of respective target scores $\{q_k^\mathcal{R}\}$.
This mapping is subsequently used during inference, to define a piecewise linear transformation between original scores and target scores.

\subsection{Application during inference}

To map original scores to transformed scores at inference time, for a given original score $s$ we find $i$ such that $q^\mathcal{S}_i \le s < q^\mathcal{S}_{i+1}$, which can be computed in $O(\log N)$ via binary search; source quantiles must be strictly increasing from $0$ to $1$, so repeated quantiles are not deployed, a score equal to a source quantile maps directly to its target, and scores outside the fitted range are clipped. Then, using the mapping discussed in the previous section we retrieve the respective target scores $q_i^\mathcal{R}$ and $q_{i+1}^\mathcal{R}$, and we transform the score as follows:

\begin{equation}
    \mathcal{T}^Q(s) = q_i^\mathcal{R} + (s - q^\mathcal{S}_i) \cdot \frac{q_{i+1}^\mathcal{R} - q_i^\mathcal{R}}{q^\mathcal{S}_{i+1} - q^\mathcal{S}_i} \ .
\end{equation}

\subsection{Example}

For the fraud detection applications, the highest quantiles are more important, because fraud events are rare. For example, we may consider quantiles $[q_1, q_2, \dots, q_6] = [0.0, 0.5, 0.95, 0.99, 0.999, 1.0]$ to be mapped to target scores $[q^\mathcal{R}_1, q^\mathcal{R}_2, \dots, q^\mathcal{R}_6] = [0, 0.2, 0.4, 0.6, 0.8, 1]$.

Without loss of generality, let's assume the original score distribution follows a $\text{Beta}(0.4, 0.7)$, meaning the application of equation~\ref{quantile_map_eq} results in $[q^\mathcal{S}_1, q^\mathcal{S}_2, \dots, q^\mathcal{S}_6] = [0, 0.268, 0.960, 0.996, 0.9998, 1]$ . Consider a score of $s=0.97$. First, we find the index i such that $q^\mathcal{S}_i \leq s \leq q^\mathcal{S}_{i+1}$.
In this case, $i=3$ as $q_3 = 0.960 \leq 0.97 \leq 0.996 = q_4$.
Then, we apply the formula from Equation~\ref{eq:quantile_transformation_linear} to find the transformed score value.
\[
\mathcal{T}^Q(0.97) = 0.4 + (0.97-0.960) \cdot \frac{0.6-0.4}{0.996-0.960} \approx 0.455
\]
Thus, the $\mathcal{T}^Q$ transforms the raw score of $0.97$ to a calibrated target score of $\approx 0.455$.

\section{Per-Client Breakdown of $p_{\text{custom}}$}
\label{sec:extra_results}

\begin{table}
    \centering
    \small
    \begin{tabular}{lrrrrr}
        \toprule
        \textbf{Bin} & \textbf{A (UC 1)} & \textbf{B} & \textbf{A (UC 3)} & \textbf{C} & \textbf{D} \\
        \midrule
        {[0,.1[} & 1 & $-4$ & 3 & 1 & $-1$ \\
        {[.1,.2[} & 0 & 3 & $-2$ & 2 & 0 \\
        {[.2,.3[} & $-2$ & 6 & $-5$ & 2 & 13 \\
        {[.3,.4[} & 5 & 14 & $-5$ & $-6$ & $-2$ \\
        {[.4,.5[} & 8 & 0 & $-12$ & $-18$ & $-32$ \\
        {[.5,.6[} & 2 & $-3$ & $-16$ & $-16$ & 82 \\
        {[.6,.7[} & 9 & 4 & $-12$ & $-3$ & $-22$ \\
        {[.7,.8[} & 13 & $-1$ & $-6$ & $-4$ & $-44$ \\
        {[.8,.9[} & $-2$ & $-16$ & $-13$ & $-13$ & $-100$ \\
        {[.9,1]} & $-33$ & $-26$ & $-26$ & $-11$ & 708 \\
        \bottomrule
    \end{tabular}
    \caption{Median relative error (\%) of $p_{\text{custom}}$ per score bin and client (UC: use case).}
    \label{tab:per_client_bins}
\end{table}

Table~\ref{tab:per_client_bins} breaks down the relative error of $p_{\text{custom}}$ in Figure~\ref{fig:transformation_update_error} by client and score bin. Each cell is the median over the client's evaluation windows (Table~\ref{tab:client_metrics}). For clients A, B and C, the median error stays within $\pm 18\%$ up to $0.8$. Client D, with two evaluation windows and $\approx$4.3k events per week, expects only a handful of events per week in the upper bins, so its errors in those bins reflect sampling noise.

\end{document}